\documentclass{article}

\usepackage{microtype}
\usepackage{graphicx}
\usepackage{subfigure}
\usepackage{booktabs} 

\usepackage{amsmath}
\usepackage{etoolbox}

\usepackage{amssymb}
\usepackage{algorithmic}
\usepackage{forloop}
\usepackage{fancyhdr}
\usepackage{stmaryrd}
\usepackage{textcomp}
\usepackage{bm}

\usepackage{tabularx}
\usepackage{multirow}
\usepackage{booktabs}
\usepackage{boldline}
\usepackage{amssymb}
\usepackage{bbm}
\usepackage{romannum}

\usepackage{amssymb}
\usepackage{booktabs}
\usepackage{multirow}
\usepackage{kotex}
\usepackage{tablefootnote}
\usepackage{footnote}
\usepackage[dvipsnames,table,xcdraw]{xcolor}
\usepackage{hyperref}
\usepackage{graphicx}
\usepackage{ulem} 
\usepackage{textcomp}
\usepackage{latexsym}
\usepackage{graphicx}
\usepackage{tabularx}
\usepackage{graphics}
\usepackage{adjustbox}
\usepackage{algorithmic}
\graphicspath{{./figures/}}

\usepackage{relsize}
\usepackage{mathtools}
\usepackage{hyperref}

\usepackage[accepted]{icml2021}

\icmltitlerunning{Robust face anti-spoofing framework with Convolutional Vision Transformer}

\begin{document}

\twocolumn[
\icmltitle{Robust face anti-spoofing framework with Convolutional Vision Transformer}



\icmlsetsymbol{equal}{*}

\begin{icmlauthorlist}
\icmlauthor{Yunseung Lee}{KakaoBank}
\icmlauthor{Youngjun Kwak}{KakaoBank,KAIST}
\icmlauthor{Jinho Shin}{KakaoBank}
\end{icmlauthorlist}

\icmlaffiliation{KakaoBank}{Division of Research and Development, KakaoBank Corp., Republic of Korea}
\icmlaffiliation{KAIST}{Department of Electrical Engineering, KAIST, Republic of Korea}

\icmlcorrespondingauthor{Jinho Shin}{william.shin@lab.kakaobank.com}

\icmlkeywords{Face Anti-Spoofing, Domain Generalization, Convolutional Vision Transformer, Global and Local Features, Hybrid Feature Extraction}

\vskip 0.3in ]



\printAffiliationsAndNotice{}  

\begin{abstract}
Owing to the advances in image processing technology and large-scale datasets, companies have implemented facial authentication processes, thereby stimulating increased focus on face anti-spoofing (FAS) against realistic presentation attacks. Recently, various attempts have been made to improve face recognition performance using both global and local learning on face images; however, to the best of our knowledge, this is the first study to investigate whether the robustness of FAS against domain shifts is improved by considering global information and local cues in face images captured using self-attention and convolutional layers. This study proposes a convolutional vision transformer-based framework that achieves robust performance for various unseen domain data. Our model resulted in 7.3\%$p$ and 12.9\%$p$ increases in FAS performance compared to models using only a convolutional neural network or vision transformer, respectively. It also shows the highest average rank in sub-protocols of cross-dataset setting over the other nine benchmark models for domain generalization.
\end{abstract}

\section{Introduction}
\label{sec:intro}
\begin{figure}[t!] 
\label{fig:concept}
\begin{center}
\includegraphics[width=\columnwidth]{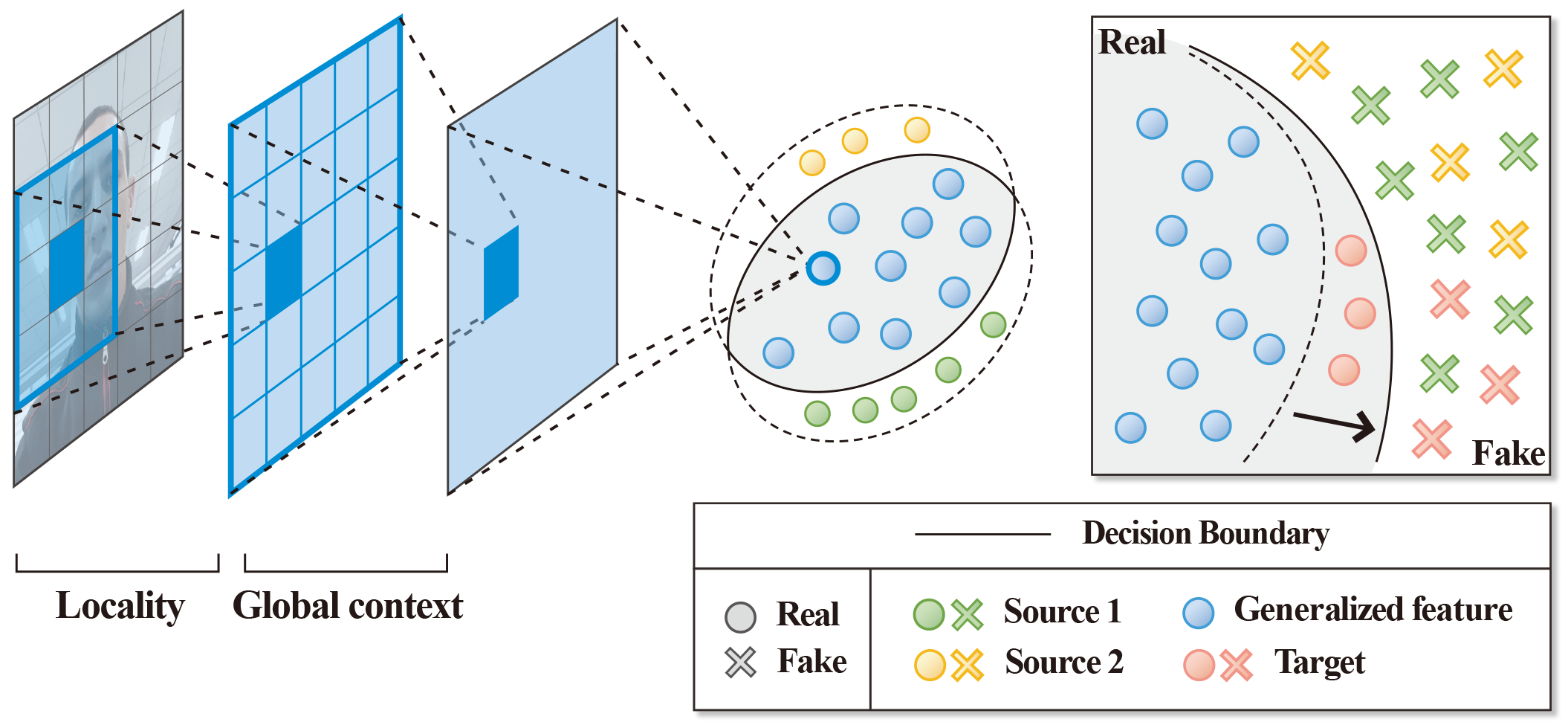}
\end{center}
\caption{The main idea of our framework for FAS comprises a hybrid feature extraction and training mechanism with a regression-based DG method. These modules lead our framework to capture more generalized feature space and to achieve robust performance on unseen target data despite distribution shifts between source and target data.}
\end{figure} 

Face anti-spoofing (FAS) aims to detect attacks that occur during face recognition. This is essential for the security of facial authentication systems because it is the first stage of the process \cite{anthony2021review}. Recently, as more diverse and sophisticated presentation attacks, such as printed images and video replay, threaten the system, the need for robust FAS algorithms is increasing \cite{yu2022deep}.
It is essential to develop a model that considers scenarios where the distribution of training data is distinct from that of data flowing into a real-time system to ensure the FAS model performs optimally in real service.
Therefore, a training mechanism based on domain generalization (DG) should be considered \cite{anthony2021review,yu2022deep,shao2019multi,jia2020single, liu2021adaptive, wang2022domain}. 
Consistent with previous studies, our study aims to develop a generalized FAS model to handle domain shifts. To achieve the robustness of FAS, we designed a framework to extract rich information from images and learn a generalized feature space that is irrelevant to the domain- and person-specific attributes, as shown in Figure \ref{fig:concept}.

Maximizing the extraction of meaningful information from an image leads to better performance in a range of vision tasks, such as FAS.
Therefore, research has focused on identifying new backbones to learn high-quality features. For example, convolution neural networks (CNNs), which are widely used as a representative backbone for FAS are advantageous, in extracting local information \cite{nagpal2019performance}. Recently, vision transformer (ViT), designed to encode the global context between image patches based on the self-attention (SA) mechanism, has emerged as a new backbone for facial authentication \cite{george2021effectiveness, huang2022adaptive, ming2022vitranspad}. 
Despite a research report indicating improved FAS performance when utilizing both local and global information in an intra-dataset protocol \cite{wang2022conv}, empirical studies on this are scarce, and prior studies have not confirmed the performance in terms of DG.

\begin{figure*}[t!] 
\begin{center}
\includegraphics[width=\linewidth]{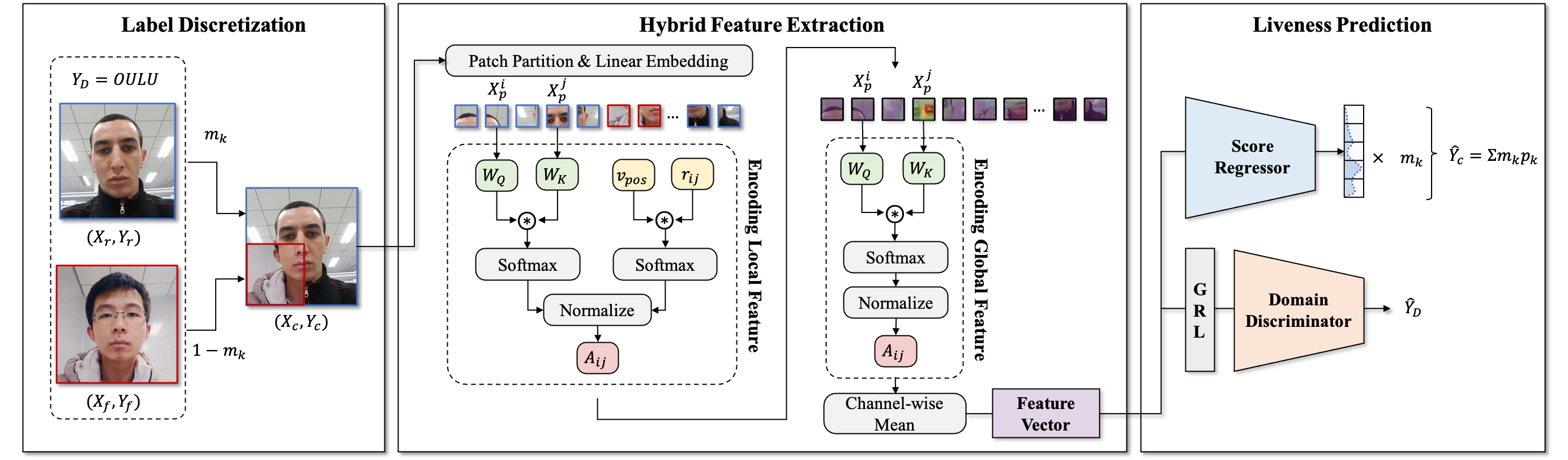}
\end{center}
\caption{
A robust FAS framework based on ConViT aims to extract local and global features with gated positional self-attention mechanisms and to learn feature space for domain generalization through liveness prediction with discretized labels.
}

\label{fig:framework}
\end{figure*} 

Few studies have focused on unknown attacks that partially cover domain shifts, with $N$-shot learning \cite{george2021effectiveness, huang2022adaptive}. Our research differs from these studies in that they only use global dependencies and overlook local information. As an example of FAS for video, there is a study that considers the effectiveness of feature extraction for DG and utilizes SA in ViT to learn spatio-temporal relationships in videos with frames \cite{ming2022vitranspad}. However, no attempt in FAS for images has been made for feature extraction to encode both local cues and global dependencies with ViT architecture.

The novelty of this study lies in bridging the gap between research on improving FAS performance with global and local information and maintaining the robustness of FAS with domain-generalized features. 
To the best of our knowledge, this is the first study to propose a robust FAS model in terms of DG using ViT structure that encodes both global and local contexts from a single image.
To make our framework sufficiently robust for DG, we designed it to extract rich features using convolutional vision transformer (ConViT) and to train it with a regression-based method as well as adversarial learning. 
Under cross-dataset protocol, our proposed framework reports a higher Area Under Curve (AUC) than single-operation-based models, such as EfficientNet and ViT. 
Our method achieved the highest performance out of nine FAS models in terms of DG, demonstrating that a rich representation trained with the generalization method contributes to the robustness of domain shifts.

\section{Related Work}

\subsection{Feature Extraction for FAS}
Depending on the feature extraction method used in neural networks, the backbones for FAS can be classified as CNN, SA-based ViT, and hybrid ViT with convolutional elements \cite{han2022survey}.

CNN-based models can extract local features from images and are trained efficiently based on strong assumptions, such as spatial equivariance. For example, ResNet \cite{he2016deep, shao2019multi, jia2020single} and EfficientNet \cite{tan2019efficientnet, kong2022detect} have been widely used to obtain local cues from images in FAS. However, the information from CNN is limited to the range of the receptive field, and it is difficult to aggregate information from other spatial locations \cite{raghu2021vision}. The performance ceiling is unavoidable if the assumption of locality and invariance does not fit the task because the CNN is operated based on complex assumptions \cite{d2021convit}.

To overcome the limitations of CNNs, the ViT-based model designs the SA mechanism with a soft assumption. ViT divides an image into non-overlapped patches and transforms them into a linearly embedded sequence. Global dependencies between patches are encoded via SA \cite{dosovitskiy2020image, liu2021swin}. However, it is difficult to solve the FAS task using ViTs because a large dataset is necessary for pretraining ViT owing to a lack of prior knowledge \cite{raghu2021vision,d2021convit,tu2022maxvit}.

Therefore, hybrid models have recently been proposed to reflect the advantages of CNNs in ViT \cite{dosovitskiy2020image,d2021convit,tu2022maxvit}. For example, ViT Hybrid uses a feature map extracted through a CNN as an input to ViT. ConViT indirectly reflects locality into the ViT structure by adding a modified SA module that operates similarly to the convolution in the initial stage \cite{d2021convit}. In addition, MaxViT is an advanced model of CoAtNet \cite{dai2021coatnet} that uses convolution layers directly in several stages of ViT \cite{tu2022maxvit}.

\subsection{Domain Generalization}
The diversity of face images inputted into the authentication system is attributed to varying environmental conditions, such as background and camera angle, between individuals. This variation causes distribution shifts between the training and inference data, which degrades model performance. Therefore, the previous studies have suggested several methods to train the generalized feature space underlying some source domains and unseen but related target domains \cite{shao2019multi, jia2020single, liu2021adaptive, wang2022domain, ming2022vitranspad}.

These studies have defined FAS as a binary classification problem that discriminates between real and spoof and adopted binary cross-entropy (BCE) as the loss function. However, recent studies have indicated that models trained via BCE are prone to attacks due to overfitting, thereby leading to decreased generalization performance. To address this issue, previous studies redefined the FAS task as a regression problem and probabilistically estimated its liveness score \cite{jiang2022ordinal, kwak2023liveness}. Consequently, the regression-based FAS model improves the generalization ability because the converted continuous label contains more semantic information from both real and spoof images \cite{jiang2022ordinal}. In addition, the expected liveness score-based regression neural network using the pseudo-discrete label encoding technique contributes to more robustness in unseen datasets than the model trained with BCE \cite{kwak2023liveness}.

\section{Proposed Method}
In this study, we propose a FAS framework with SA and a convolution-based feature extractor for DG. Figure \ref{fig:framework} shows our proposed framework consists of three stages: label discretization, feature extraction, and liveness prediction.  

\subsection{Label Discretization}
We redefine the binary FAS classification as a regression-based method, as this approach is more suitable for learning generalized and discriminative representations \cite{jiang2022ordinal,kwak2023liveness}. A sequence of class labels was transformed into a probability distribution using CutMix \cite{yun2019cutmix}. CutMix is a technique used for data augmentation, but we generate a label with CutMix to discretize the pseudo-label and solve the FAS task with a regression-based loss function. Specifically, there are $N$ source domain datasets, which comprise face images $X \in \mathbb{R}^{H{\times}W{\times}3}$ with binary label $Y {\in} {\{0,1\}}$ that represents fake and real for each face image. The real and fake face images are denoted as $X_r$ and $X_f$, respectively, and $Y_r$ and $Y_f$ are the corresponding labels. Using CutMix, we swap some parts of $X_r$ to that of $X_f$ and denote this combined image as input image $X_c$. We define the discretized pseudo-label $Y_c$ as $m_lY_r$, which is the single value $m_l$ selected from the set $M=\{0,\frac{1}{K},\frac{2}{K},…,\frac{K-1}{K},1\}$, where the interval $[0,1]$ is divided by a constant $K$.

\subsection{Hybrid Feature Extraction}
In the second stage, we employ ConViT as a backbone $F$ for feature extraction to encode rich information to capture a cue for FAS. Specifically, global context and local cues are obtained through gated positional self-attention(GPSA) which is modified from SA. 
In general, an attention score A in original ViT is defined as the dot product of query embedding $Q$ and key embedding $K$. $W_Q$, $W_K$, and $W_V$ are the weight matrices for $Q$, $K$, and $V$, respectively, which are embeddings for the query, key, and value, respectively. The embeddings were inferred by multiplying the linearly projected input $X_p$ with the corresponding weight matrices. The attention score $A_{ij}$ implies the semantic relevance between patches $X_p^i$ and $X_p^j$ and can capture the long-range dependency between patches. 

To reflect locality in the attention score, an inner product of the learnable embedding $v_{pos}$ and relative positional encoding $r_{ij}$ is added to $Q_iK_j^T$. As shown in Eq. \eqref{eq:3}, the gating parameter $\sigma$ manipulates the type of information on which to focus. A larger $\sigma$ at the initial stage makes this GPSA function as generalized convolution, which assigns higher weights to adjacent patches. After extracting the local features at the earlier layer, a smaller $\sigma$ is gated with more global information of patches far from each other at the upper layer.
\begin{equation}
GPSA(X_p) := normalized[A]X_pW_v,
\label{eq:2}
\end{equation}
\begin{equation} \label{eq:3}
\begin{split}
where \ A_{ij} = (1-\sigma)\times softmax(Q_iK_j^T) \\ +\sigma \times softmax(v_{pos}^Tr_{ij}).
\end{split}
\end{equation}

\subsection{Liveness Prediction}
The feature vector is derived from the input image $X_c$, by averaging the patch embedding $F(X_c)$ extracted from our backbone $F$ and passed to both the score regressor $R$ and domain discriminator $D$, which contributes to a robust backbone for unseen domain data \cite{jia2020single}. As described in Eq. \eqref{eq:4}, $p_k^i$ is the probability vector of the real image in the $i^{th}$ input image, and the liveness score $\hat{Y}_c^i$ is calculated using the sum of the element-wise multiplication between $M$ and corresponding probabilities in $p_k^i$. The score regressor $R$ is trained to minimize the difference between the predicted score $\hat{Y}_c^i$ and groundtruth label $Y_c^i$ through the mean squared error. Subsequently, backbone $F$ is trained adversarially to extract generalized features, which makes it difficult for the discriminator to predict the domain label $Y_D$ of a feature. For adversarial learning, a gradient reversal layer (GRL) is included after the feature extractor to make it difficult for the discriminator to distinguish features from different domains. Additionally, the regressor and discriminator are composed of fully connected layers. In Eq. \eqref{eq:5}, $q(\cdot)$ is the probability vector of the discriminator and is equal to $D(F(X_c))$. Therefore, domain-invariant feature is extracted from backbone $F$ through adversarial learning, as described in Eq. \eqref{eq:5}. The final loss function is defined as in Eq. \eqref{eq:6}. 
\vspace{-0.1cm}
\begin{equation}\label{eq:4}
\mathcal{L}_{reg}=\sum \Vert Y_c^i-\hat{Y}_c^i\Vert_2^2, \\ \quad where \; \hat{Y}_c^i = \sum_{k=0}^Km_k\times p_k,
\end{equation}
\begin{equation} \label{eq:5}
\min_{D}\max_{F}\mathcal{L}_{adv}(F,D)=-\mathbb{E}_{x,y{\sim}{X_c, Y_D}}\sum_{n=1}^N \mathbbm{1}_{[n=y]}\log{q(x)},
\end{equation}
\begin{equation}
\mathcal{L}_{final} = \mathcal{L}_{reg} + \mathcal{L}_{adv}.
\label{eq:6}
\end{equation}

\section{Experimental Results}
\subsection{Experimental Settings}
We conducted experiments on various FAS benchmark datasets, namely OULU-NPU(O) \cite{boulkenafet2017oulu}, MSFD-MSU(M) \cite{wen2015face}, REPLAY-ATTACK(I) \cite{chingovska2012effectiveness}, and CASIA-FASD(C) \cite{zhang2012face}. 
We have used a cross-dataset protocol to evaluate the performance. The cross-dataset protocol is a leave-one-out setting that is used to train a model with several source datasets and to test the generalization performance with an unseen target dataset. Half total error rate (HTER) and area under curve (AUC) were used as evaluation metrics \cite{yu2022deep}.

We used CNN, ViT, and hybrid networks pretrained with ImageNet1K to prevent overfitting and achieve further improvement in DG \cite{parkin2019recognizing, ming2022vitranspad}. While we compared the FAS performance depending on differences in the feature extraction methods of the backbones, the structure of the score regressor and discriminator was kept the identical, and each feature extractor was trained with the same loss function as in Eq. \eqref{eq:6}. 

\subsection{Experimental Results}
\begin{table}[]
\renewcommand{\arraystretch}{1.15}
\caption{Results of the cross-dataset protocol. In each cell, the left and right numbers represent HTER(\%) and AUC(\%), respectively. \textbf{Bold} implies the best results and second best is \underline{underlined}.}
\vspace{0.1cm}
\resizebox{\columnwidth}{!}{%
\begin{tabular}{c|c|c|c|c|c||c}
\hline
\multicolumn{2}{c|}{\textbf{Backbone}}                           & \textbf{OCI$\rightarrow$M} & \textbf{OMI$\rightarrow$C} & \textbf{CMO$\rightarrow$I} & \textbf{CMI$\rightarrow$O} & \textbf{$\mu \pm \sigma$}     \\ \hline
\multirow{2}{*}{\textbf{CNN}}    & \textbf{ResNet}     & 5.8 / 91.8     & 14.9 / 89.4    & \textbf{11.5 /   94.4}  & 15.0   / 90.5  & \underline{11.8}+4.3   / 91.5+2.2 \\ \cline{2-7} 
 & \textbf{EfficientNet} & 10.4 / 92.4 & 18.8 / 84.2   & 18.9 / 87.4 & 21.7 / 82.5    & 17.5+4.9 /   86.6+4.3 \\ \hline
\multirow{2}{*}{\textbf{ViT}}    & \textbf{ViT Base}   & 17.9 / 83.7    & 31.3 / 70.7    & 16.9 / 89.9    & 28.3 / 79.5    & 23.6+7.3 /   81.0+8.1 \\ \cline{2-7} 
 & \textbf{Swin Transformer}     & 10.0 / 94.3 & 23.7 / 77.0   & 16.0 / 78.5 & \textbf{10.6 /   95.2} & 15.1+6.3 /   86.2+9.8 \\ \hline
\multirow{3}{*}{\textbf{Hybrid}} & \textbf{ViT Hybrid} & \textbf{5.0} / \underline{97.1}     & \underline{13.6} / \underline{92.2}    & 17.9 / 89.9    & \underline{11.3} / \underline{95.0}    & 12.0+5.4 /   \underline{93.5}+3.1 \\ \cline{2-7} 
 & \textbf{ConViT}       & \textbf{5.0 / 97.2}  & \textbf{12.1 /   93.6} & \underline{15.1} / \underline{93.3} & 14.7 / 91.6    & \textbf{11.7}+4.7   / \textbf{93.9}+2.4 \\ \cline{2-7} 
 & \textbf{MaxViT}       & 10.4 / 93.9 & 14.5 / 89.7   & 26.8 / 72.9 & 20.6 / 87.1    & 18.1+7.1 /   85.9+9.1 \\ \hline
\end{tabular}%
}\label{tab:cross}
\end{table}

We conducted a comparative experiment to determine the most robust FAS backbone among the candidate models. Specifically, we selected ResNet and EfficientNet as representative models for the CNN, ViT and swin transformer for the ViT series and ViT Hybrid, ConViT, and MaxViT for the hybrid series. 
First, Table \ref{tab:cross} demonstrates that main feature extraction operation significantly affects the performance. 
When ConViT was used as a feature extractor, it achieved the best performance with an HTER of 11.7\% and AUC of 93.9\% compared to the CNN and ViT models. In addition to ConViT, ViT Hybrid had the second-highest average AUC. However, the ViT and swin transformer using only SA exhibit HTER scores of 23.6\% and 15.1\%, respectively, recording the highest error rates, which implies that the ViT is vulnerable to changes in the distribution of the evaluation data. When features were extracted from a ViT-based hybrid model containing convolution-like elements, performance in the unseen domain was typically superior to that of convolution or SA models alone. Specifically, the ConViT-based framework outperformed EfficientNet and ViT in 7.3\%$p$ (=93.9\%-86.6\%) and 12.9\%$p$ (=93.9\%-81.0\%) higher AUC scores, respectively.

Second, we found that convolution-like elements in the initial layer would be effective in improving FAS performance by comparing the hybrid models, as shown in Table \ref{tab:cross}. Specifically, MaxViT using convolution in most stages demonstrated poor performance, while ConViT, which indirectly adds convolutional SA in the early stage, and ViT Hybrid, which utilizes convolution in the input image, reported the lowest HTER values among the comparison groups. In addition, both MaxViT and swin transformer are similar in that they use window-based SA and the averaged AUC score of these models is about 86\% compared to that of the other hybrid models, which are ConViT and ViT Hybrid, achieving 94\%. Therefore, models using window-based SA show an approximately 8\%$p$ lower AUC than the other hybrid models. Furthermore, we experimentally confirm that attention using a window shift is inappropriate for FAS; however, convolutional elements in an early stage of the network contribute to robust classification.

\begin{table}[]
\caption{Comparison with other generalized FAS methods under cross-dataset testing. Results are evaluated as average ranks in each setting. $\dagger$ and $\ast$ denote the proposed framework with ViT Hybrid and ConViT, respectively.}
\vspace{0.1cm}
\resizebox{\columnwidth}{!}{%
\begin{tabular}{l|c|c|c|c|c}
\hline
\textbf{Model}               & \textbf{OCI$\rightarrow$M} & \textbf{OMI$\rightarrow$C} & \textbf{CMO$\rightarrow$I} & \textbf{CMI$\rightarrow$O} & \textbf{Avg. Rank} \\ \hline
\textbf{MADDG} \cite{shao2019multi}      & 17.7 / 88.1   & 24.5 / 84.5 & 22.2 / 85.0 & 27.9 / 80.0 & 10.75 \\ \hline
\textbf{PAD-GAN} \cite{wang2020cross}    & 17.0 / 90.1   & 19.7 / 87.4 & 20.9 / 86.7 & 25.0 / 81.5 & 8.75  \\ \hline
\textbf{RF-Meta} \cite{shao2020regularized}    & 13.9 / 94.0   & 20.3 / 88.2 & 17.3 / 90.5 & 16.5 / 91.2 & 6.5   \\ \hline
\textbf{SSDG-M} \cite{jia2020single}     & 16.7 / 90.5   & 23.1 / 85.5 & 18.2 / 94.6 & 25.2 / 81.8 & 9     \\ \hline
\textbf{SDA} \cite{wang2021self}        & 15.4 / 91.8   & 24.5 / 84.4 & 15.6 / 90.1 & 23.1 / 84.3 & 7     \\ \hline
\textbf{$\textbf{D}^2$AM} \cite{chen2021generalizable}                & 15.4 / 91.2   & \underline{12.7} / \textbf{95.7} & 21.0 / 85.6 & 15.3 / 90.9 & 5.75  \\ \hline
\textbf{ANRL} \cite{liu2021adaptive}      & 10.8 / 96. 8  & 17.8 / 89.3 & 16.0 / 91.0 & 15.7 / \underline{91.9} & 4.5   \\ \hline
\textbf{SSAN-M} \cite{wang2022domain}     & 10.4 / 94.8         & 16.5 / 90.8          & \textbf{14.0 / 94.6} & 19.5 / 88.2          & 4                  \\ \hline
\textbf{ViTransPAD} \cite{ming2022vitranspad} & \underline{8.4} / - & 17.9 / -    & 16.0 / -    & 15.7 / -    & 4.25  \\ \hline \hline
\textbf{Ours$\dagger$} & \textbf{5.0} / \underline{97.1} & 13.6 / 92.2          & 17.9 / 89.9          & \textbf{11.3 / 95.0} & \underline{3}            \\ \hline
\textbf{Ours$\ast$}      & \textbf{5.0 / 97.2} & \textbf{12.1} / \underline{93.6} & \underline{15.1} / \underline{93.3}    & \underline{14.7} / 91.6    & \textbf{1.5}       \\ \hline
\end{tabular}%
}\label{tab:other}
\end{table}

Table \ref{tab:other} presents a comparison of the proposed framework with the latest promising models. The ConViT or ViT Hybrid-based framework was ranked 1.5 and 3, respectively, by averaging HTER scores over four sub-protocols in a cross-dataset setting, showing the highest performance.
In particular, the proposed ConViT-based model outperformed a previous model with ViT structure \cite{ming2022vitranspad}, exhibiting a 2.8\%$p$ lower HTER. These results suggest that our ConViT-based model is the most effective as a generalized FAS methodology compared to the previously developed methodologies \cite{shao2019multi,wang2020cross,shao2020regularized,jia2020single,wang2021self,chen2021generalizable,liu2021adaptive,wang2022domain,ming2022vitranspad}, and the extraction of rich representations from images significantly impacts improved performance in comparison to learning the spatio-temporal relationship in a video \cite{ming2022vitranspad}. Consequently, the results suggest that both locality and global dependencies within patches contribute to a more robust performance.

\section{Conclusion}
This study demonstrates that our hybrid FAS framework with self-attention and convolution is more robust for FAS tasks than those with a CNN and ViT alone. Our ConViT-based framework improved the performance of domain generalization, over other promising FAS models. These results suggest that it is important to extract both local and global information from images for a robust FAS against domain shifts.


\bibliography{main}
\bibliographystyle{icml2021}
\clearpage

\end{document}